\documentclass[letterpaper, 10 pt, conference]{ieeeconf}  

\IEEEoverridecommandlockouts                              

\usepackage{amsmath} 
\usepackage{graphicx} 
\usepackage{hyperref}
\usepackage{siunitx}

\usepackage{booktabs}
\usepackage{multirow}
\usepackage{tikz}

\title{\LARGE \bf
A Map-Free LiDAR-Based System for Autonomous Navigation in Vineyards}

\author{Riccardo Bertoglio$^{1}$, Veronica Carini$^{1}$, Stefano Arrigoni$^{2}$, and Matteo Matteucci$^{1}$
\thanks{$^{1}$Department of Electronics, Information and Bioengineering,
            Politecnico di Milano, Milan, Italy
            {\tt\small \{name.surname\}@polimi.it}}%
\thanks{$^{2}$Department of Mechanical Engineering,
            Politecnico di Milano, Milan, Italy
            {\tt\small stefano.arrigoni@polimi.it}}
}

\newcommand\copyrighttext{%
  \footnotesize 
\textcopyright2023 IEEE.  Personal use of this material is permitted.  Permission from IEEE must be obtained for all other uses, in any current or future media, including reprinting/republishing this material for advertising or promotional purposes, creating new collective works, for resale or redistribution to servers or lists, or reuse of any copyrighted component of this work in other works.}
\newcommand\copyrightnotice{%
\begin{tikzpicture}[remember picture,overlay]
\node[anchor=south,yshift=10pt] at (current page.south) {\fbox{\parbox{\dimexpr\textwidth-\fboxsep-\fboxrule\relax}{\copyrighttext}}};
\end{tikzpicture}%
}

\begin{document}

\maketitle
\copyrightnotice
\thispagestyle{empty}
\pagestyle{empty}

\begin{abstract}

Agricultural robots have the potential to increase production yields and reduce costs by performing repetitive and time-consuming tasks. However, for robots to be effective, they must be able to navigate autonomously in fields or orchards without human intervention. In this paper, we introduce a navigation system that utilizes LiDAR and wheel encoder sensors for in-row, turn, and end-row navigation in row structured agricultural environments, such as vineyards. Our approach exploits the simple and precise geometrical structure of plants organized in parallel rows. We tested our system in both simulated and real environments, and the results demonstrate the effectiveness of our approach in achieving accurate and robust navigation. Our navigation system achieves mean displacement errors from the center line of \SI{0.049}{\metre} and \SI{0.372}{\metre} for in-row navigation in the simulated and real environments, respectively. In addition, we developed an end-row points detection that allows end-row navigation in vineyards, a task often ignored by most works.

\end{abstract}

\section{Introduction}

The increasing demand for food in the current climate-changing environment introduces new challenges, such as the necessity of increasing production and the sustainability of crop management while reducing costs~\cite{bertoglio2021digital}. Agricultural robots can help achieve these goals by performing repetitive and time-consuming tasks, allowing farmers to improve production yields. At the same time, for robots to be effective, they must be able to navigate autonomously in fields or orchards without human intervention. Navigation approaches can be broadly divided into two categories: those with or without a map of the environment. While map-based approaches can be helpful in unstructured environments, they require a more expensive sensor suite and incur increased computational effort. Additionally, localization on a pre-built map can fail due to the constantly changing nature of agricultural environments. Nevertheless, agricultural environments typically have a simple and precise geometrical structure, with crops organized in parallel rows. This structure can be exploited for navigation without the need for a map. 

Autonomous navigation in agriculture often utilizes GNSS information for pre-planned routes or as supplementary information. Additionally, Differential GNSS technology provides higher localization accuracy of up to centimeters. However, the GNSS signal is not always available, especially for those cultivations with high plants and abundant vegetation. LiDAR and camera sensors are also utilized for navigation. LiDARs can be either 2D or 3D sensors, with the latter characterized by multiple scanning planes. LiDAR sensors provide a geometrical view of the environment, work at a reasonable frequency (over 10 Hz), and are precise. Cameras, such as RGB, stereo, or RGB-D, provide a more complex semantic interpretation of the environment, which is helpful for tasks like obstacle avoidance. Stereo and RGB-D cameras can also produce 3D renderings of the environment. Although LiDARs only provide geometrical data, they are less susceptible to lighting conditions than cameras, which is essential in agricultural environments where strong sunlight and shadows are typical.

\begin{figure}[t]
    \centering
    \includegraphics[width=0.6\columnwidth]{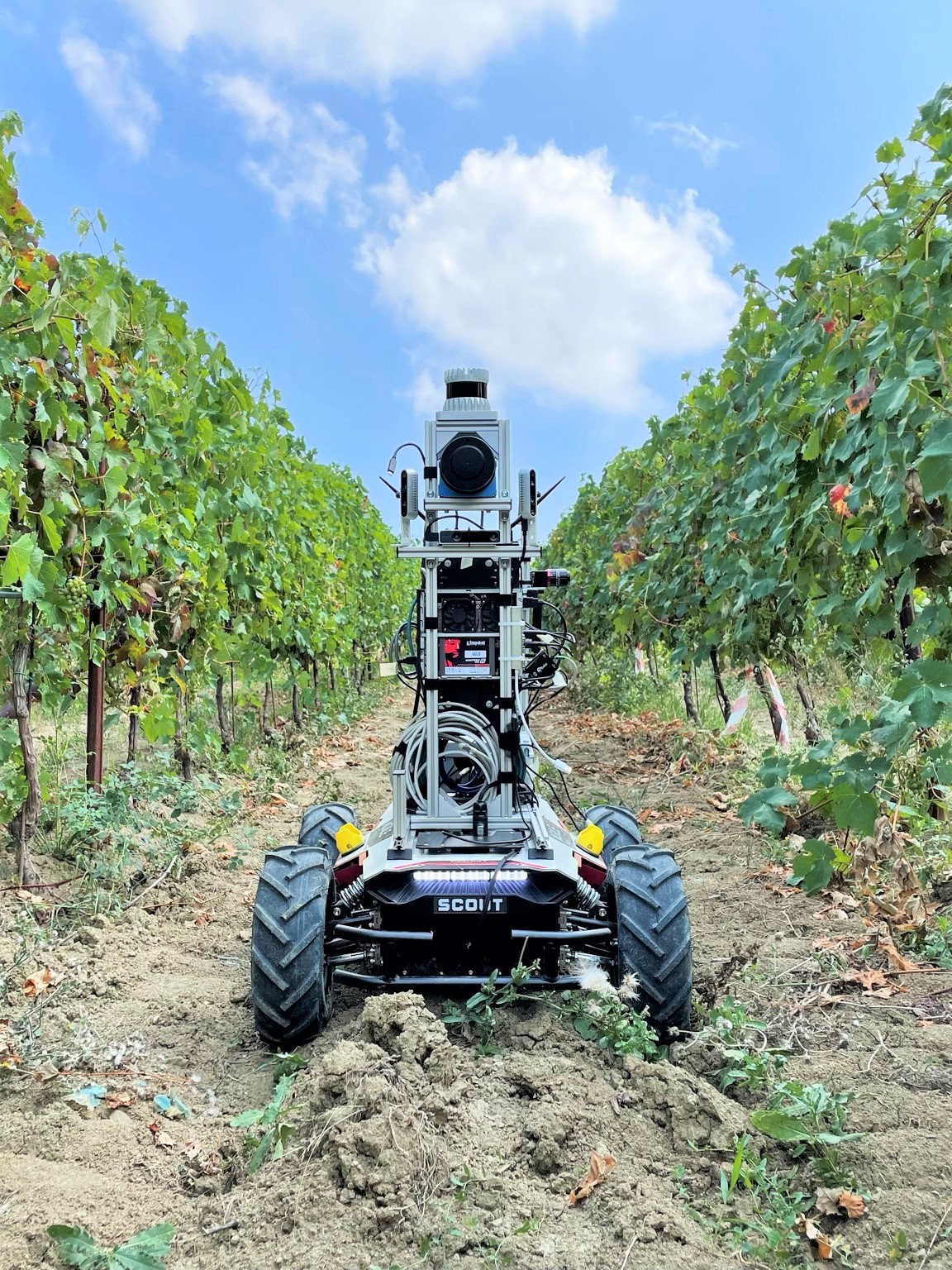}
    \caption{Our robotic platform navigating a real vineyard.}
    \label{fig:robot_vineyard}
\end{figure}

\begin{figure*}[t]
    \centering
    \includegraphics[width=0.7\textwidth]{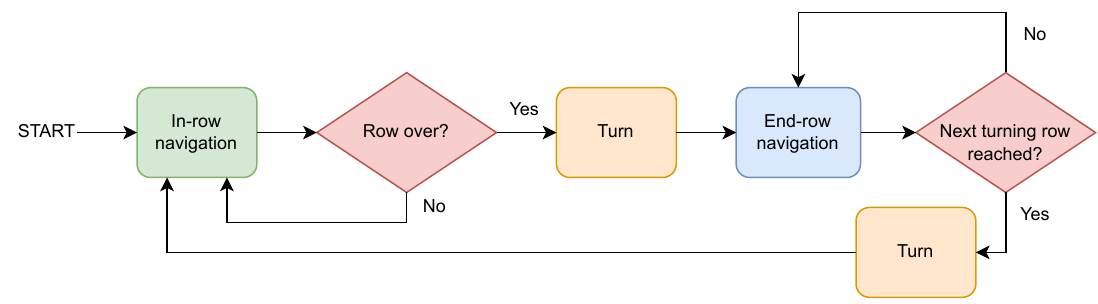}
    \caption{The general navigation software architecture.}
    \label{fig:general_arch}
\end{figure*}

The VINBOT project~\cite{guzman_aut_2016} has developed a vineyard navigation system combining a line detection algorithm and GNSS navigation for in-row navigation. Two lines representing vineyard rows were identified using a 2D laser and RANSAC algorithm. The robot changed the corridor by rotating around one of two points representing the plant's end. Localization relied on IMU, GPS, and wheel odometry data, but tests have shown that plant holes should be manually managed to avoid misinterpretation.

The VineSLAM algorithm, described in~\cite{santos_towards_2016}, employed laser rangefinder data and known parameters to identify trunks and masts as landmarks for 2D SLAM. RFID tags were utilized to mark the corridor boundaries for topological mapping. However, the algorithm's accuracy relied on the detection of trunks and masts, and external factors such as grass and wind introduced substantial noise, compromising navigation reliability.

Bernad et al.~\cite{bernad_eval_2018} proposed three straightforward in-row navigation approaches using only 2D LiDAR data. The most effective algorithm involved calculating the average distance from both sides of the crop row and estimating an orientation correction based on the offset. They achieved an accuracy of \SI[separate-uncertainty = true, multi-part-units = repeat]{0.041 \pm 0.034}{\metre} from the center line when testing outdoors with potted maize plants.

Rovira-M{\'a}s et al.~\cite{mas_augm_2021} presented a multi-sensor navigation approach for inside-row guidance. The authors used a so-called Augmented Perception Obstacle Map (APOM) to store and evaluate readings from a 3D stereo camera, LiDAR, and ultrasonic sensors. The map is then analyzed to find specific situations representing the status of row detection. The next navigation target point is only computed if one or both rows are found.

Mengoli et al.~\cite{mengoli_aut_2020, mengoli_robust_2021} proposed Hough Transform-based methods for orchard navigation, including in-row and row-change maneuvers. The authors enhanced robustness by incorporating vineyard geometry conditions and using GPS to identify corridor ends. The detected pivot point in row-change maneuvers had an RMSE of 0.3429 m in the x direction and 0.5840 m in the y direction.

Aghi et al.~\cite{aghi_local_2020} introduced a vineyard in-row navigation algorithm with two components. The first component uses an RGB-D camera's depth map to detect the end of the row by fitting a rectangular area to the farthest pixels. In case of failure, a backup algorithm takes over, utilizing a neural network to identify and correct the robot's orientation if needed.

The Field Robot Event (FRE)\footnote{\url{https://fieldrobot.nl/event}} is a robotics competition that focuses on autonomous navigation in agricultural environments. We drew inspiration from the in-row navigation approach used by the Kamaro team~\cite{kamaro2021fre} in the 2021 FRE competition for maize fields and adapted it for vineyard navigation. Our navigation system utilizes a single LiDAR and wheel encoders to reduce sensor requirements and costs. Additionally, we developed an end-row navigation algorithm to facilitate autonomous row changes. We proposed a straightforward evaluation benchmark for in-row navigation and end-row point detection, eliminating the need for external devices like laser tracking or Differential GNSS systems. The system was tested in both real vineyard (see Figure~\ref{fig:robot_vineyard}) and simulated environments. The complete algorithm code is available at this GitHub repository: \url{https://github.com/AIRLab-POLIMI/MFLB-vineyard-navigation}.

\section{Materials and Methods}

We developed our navigation algorithm for a skid-steering mobile robot, although the general structure can also be adapted to other types of kinematics. The navigation software was implemented using the Robot Operating System (ROS) library, specifically the Melodic version on Ubuntu 18.04 LTS. The software architecture is presented in Figure~\ref{fig:general_arch}. 

Initially, the robot is assumed to reach the beginning of a row; the In-row navigation module guides the robot to follow the row until the end is detected. Then, the robot performs an open-loop turn managed by the End-Row navigation module, which guides the robot along the border of the vineyard until it reaches a specified row to turn into, where the In-row navigation module is reactivated. The following gives a more detailed description of each algorithm component.

\begin{figure*}[t]
    \centering
    \includegraphics[width=0.9\textwidth]{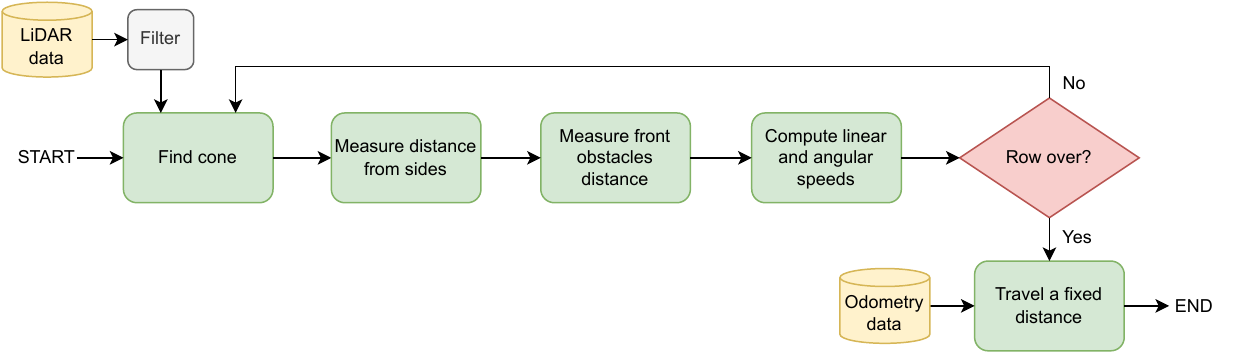}
    \caption{In-row navigation algorithm.}
    \label{fig:inrow_diagram}
\end{figure*}

\subsection{Input Data}
\label{sec:input}
Our algorithm needs very few input data, namely, an odometry source and 2D laser scans. Since we used a robot with a skid-steering kinematic, we computed its odometry with the model presented in~\cite{odometry_paper}. The kinematic relation is expressed as follows:
\begin{equation}
	\label{eq:odom1}
	\left(
 \begin{matrix}
	v_{x}  \\
	v_{y}  \\
	\omega_{z}
\end{matrix}
	\right)
	= A \cdot 
	\left(
	 \begin{matrix}
		V_{l}  \\
		V_{r}  
	\end{matrix}
    \right)
\end{equation}
where $v = (v_{x}, v_{y})$ is the vehicle’s translational velocity with respect to its local frame, $\omega_{z}$ is its angular velocity, $V_{l}$ and $V_{r}$ are the left and right linear
tread velocities, and matrix $A$ is defined by Equation \eqref{eq:odom2}. Following the experiments presented in~\cite{odometry_paper} we have calibrated the matrix A that, in the case of an ideal symmetrical kinematic, takes the following form:

\begin{equation}
	\label{eq:odom2}
	A = \frac{\alpha}{2 x_{ICR}} \cdot
	\begin{bmatrix}
		0 & 0 \\
		x_{ICR} & x_{ICR} \\
		-1 & 1
	\end{bmatrix}
\end{equation}
where, $x_{ICR}$ is the $x-axis$ component of the Instantaneous Center of Rotation (ICR), and $\alpha$ is a correction factor to account for mechanical issues such as tire inflation conditions or the transmission belt tension. Both these parameters have been empirically estimated following the directions provided in~\cite{odometry_paper}.

Beyond odometry, our navigation system expects 2D laser scans to perceive the environment. We transformed LiDAR messages from an Ouster OS1 3D LiDAR sensor into 2D laser scans through the \textit{pointcloud\_to\_laserscan} ROS package\footnote{\url{https://github.com/ros-perception/pointcloud_to_laserscan}}. We set the sensor at 10 Hz and 1024 points for each of its 64 planes. We then filtered the laser scan messages to reduce their size. We first applied radius filtering to remove points outside a circle centered on the sensor and then downsampling to reduce the density of points. We also applied outlier filtering to remove noise from data.

\subsection{In-row navigation}
In the in-row navigation stage, the navigation system makes the robot traverse a corridor created by two lines of plants by maintaining an equal distance from them as much as possible. The approach we used for the in-row navigation has been adapted from that of the Kamaro team\footnote{\url{https://github.com/Kamaro-Engineering/fre21_row_crawl}} which participated in the 2021 FRE competition.

\begin{figure}[t]
    \centering
    \includegraphics[width=0.77\columnwidth]{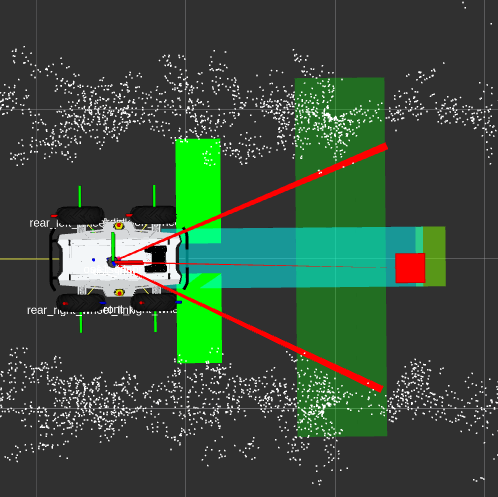}
    \caption{The robot navigating inside a row in the simulated environment. The two thick red lines represent the sides of the cone, while the red square on the center line represents the new navigation point to follow. The light green rectangles are used to compute the distances from both sides. The semi-transparent rectangle in front of the robot is used to check if the end of the row is reached by counting the number of points inside it. The rectangle placed in the middle-front part of the robot is used to check an obstacle's distance and reduce speed accordingly.}
    \label{fig:rect_cone}
\end{figure}

The functioning of the In-row navigation module is graphically illustrated in Figure~\ref{fig:inrow_diagram}. The \textit{find\_cone} method analyzes the laser scan messages to find an obstacle-free cone in front of the robot. To do so, a cone centered on the moving robot direction is gradually grown by enlarging the apex angle until a certain number of points fall inside the cone. The two cone sides are moved independently, and they have a configurable length. Once the cone is found, we compute an angular offset between the cone center line and the robot center line. This angular offset is increased by an additional offset proportional to the distance between the robot and corridor center. The latter distance is computed by growing two rectangles on the side of the robot until a certain number of points fall into them. A graphical representation of the cone and rectangles is shown in Figure~\ref{fig:rect_cone}.

The final angular offset defines a new line pointing toward the steering direction. We use a PID controller to steer toward the point on this line that is \SI{1}{\metre} in front of the robot. The linear speed is set to a constant value, and it is reduced if an object in front of the robot is detected. The algorithm uses a rectangle in front of the robot to calculate the target speed based on the distance between the robot and any obstacles.

\begin{figure*}[t]
    \centering
    \includegraphics[width=0.9\textwidth]{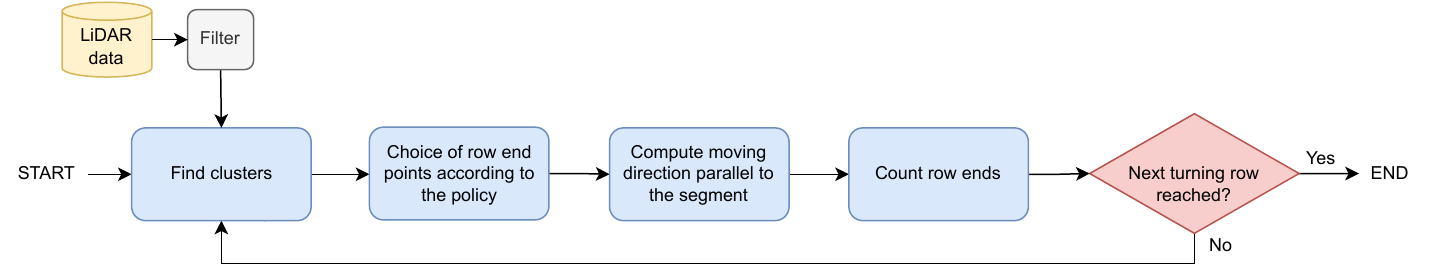}
    \caption{End-row navigation algorithm.}
    \label{fig:endrow_diagram}
\end{figure*}

At each linear and angular speed update, the In-row module checks if the end of the row has been reached. This procedure involves a rectangular area (colored light green in Figure~\ref{fig:rect_cone}) placed in front of the robot, spanning the entire corridor and part of both row sides. The corridor is over when the number of points in the rectangle approaches zero. The last step is to exit the row by a fixed distance measured through the robot odometry. Since the latter distance is usually of about \SI{1}{\metre}, the odometry guarantees a reasonable accuracy.

Once the robot has exited a row, it performs an in-place rotation by a fixed angle (usually \SI{90}{\degree}). The user needs to set the direction of the first rotation, left or right. During the rotation, the odometry is monitored to halt the robot when the required angle has been performed. Note here that we expect the robot to skid, and because of this, the effective rotation might differ from \SI{90}{\degree}. However, the algorithm overcomes this problem by selecting two end points—one positioned in front and the other at the back of the robot. Subsequently, it rotates the robot to align its moving direction parallel to the line segment connecting these two points. It's also important to note that the robot does not need to be perfectly aligned with the row direction when it begins navigating at the beginning of the row. In both scenarios, the algorithm compensates for an incomplete rotation up to a specific angle. The maximum angle that can be recovered depends on factors such as the width of the row, the robot's distance from the row's starting point, and algorithm parameters like the length of the cone sides. Once the turn is completed, the navigation system activates the End-row navigation module.

\subsection{End-row navigation}
\label{sec:endrow_navigation}

After completing the turn, the navigation system initiates the End-row algorithm. A schematic representation of the End-row navigation algorithm is presented in Figure~\ref{fig:endrow_diagram}. The primary objective of this algorithm is to enable the robot to travel perpendicularly to the field rows until it reaches the next corridor. The algorithm is specifically designed to leverage row ends, which typically consist of wooden support poles in vineyards. We employed the Euclidean Cluster Extraction technique~\cite{RusuDoctoralDissertation} to identify row ends from the 2D point cloud data. This simple algorithm is highly effective in vineyards because the rows are widely separated by open areas to allow for human operations. Each obtained cluster represents a row end.

\begin{figure}[t]
    \centering
    \includegraphics[width=0.77\columnwidth]{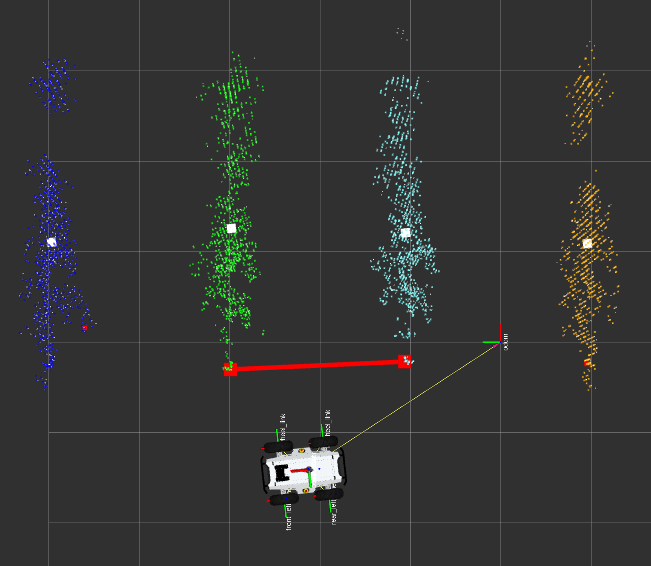}
    \caption{A screenshot of the simulation environment with the clustered row ends. Each cluster is represented with a different color. With red squares are shown the selected end points according to the Nearest policy. The red line represents the segments the robot follows to navigate perpendicularly to row ends.}
    \label{fig:clustering_segm}
\end{figure}

The subsequent task selects a point for each recognized cluster, representing the row end. We evaluated two policies to select such end point. The first policy, termed \textit{Nearest}, involves selecting the nearest cluster point to the robot center, which is surrounded by a minimum number of points at a threshold distance. Therefore, the circular neighborhood's radius and the minimum number of points are parameters that need to be configured. The second policy, called \textit{Line fitting}, involves a first step in which the end point is selected with the Nearest policy, then a line is fitted to the cluster of points, and finally, the end point is projected onto that line. We implemented line fitting using the random sample consensus (RANSAC) algorithm, finding that $100$ iterations and a distance threshold of \SI{0.1}{\metre} offer a good balance between speed and accuracy.

\begin{figure}[t]
    \centering
    \includegraphics[width=0.77\columnwidth]{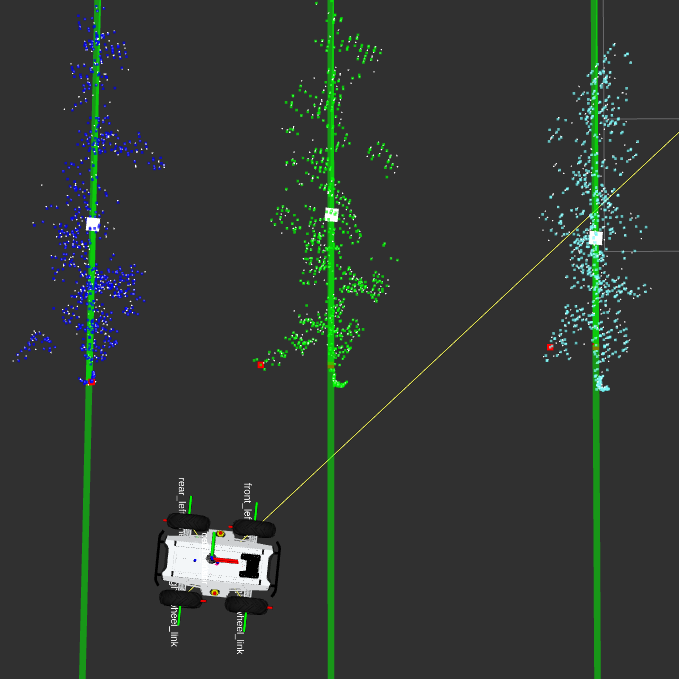}
    \caption{A screenshot of the simulation environment when the robot is performing end-row navigation. End row points are clustered, and a line is fitted for each cluster (green lines). Then, each end point (red squares) is projected onto the line model of its cluster.}
    \label{fig:line_fitting}
\end{figure}

After detecting the points representing row ends, we use them to construct segments that indicate the navigation direction. Indeed, the navigation system keeps a fixed distance from row ends by maintaining a moving direction parallel to such fitted segments. Figure~\ref{fig:clustering_segm} displays the clustered row ends in various colors and the identified end points through the Nearest policy with red squares. Additionally, the current direction segment is shown with a red line. Figure~\ref{fig:line_fitting} shows the clusters and end points obtained through the Line fitting policy.

While the robot navigates parallel to end rows, it keeps track of the number of passed row ends and stops in the middle of the next corridor to enter. Then it will perform a \SI{90}{\degree} in-place rotation, and the system will activate the In-row navigation module again.

\section{Results}

\begin{figure}[t]
    \centering
    \includegraphics[width=0.77\columnwidth]{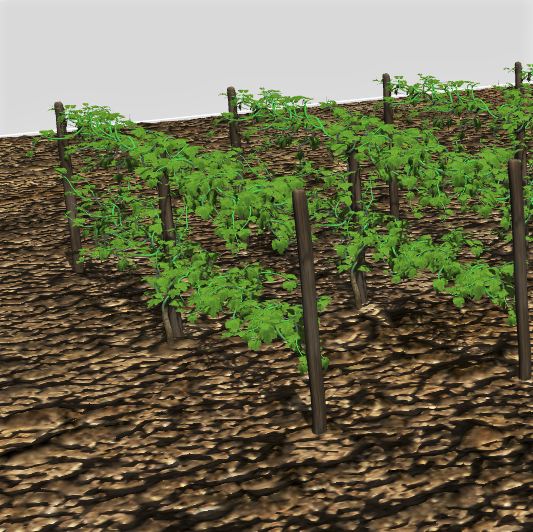}
    \caption{A screenshot that depicts a portion of the simulated vineyard.}
    \label{fig:simulation_model}
\end{figure}

We conducted experimental tests in both simulated and real environments. The simulation has been performed on the Gazebo simulator with vineyard models at different vegetative stages taken from the BACCHUS project repository\footnote{\url{https://github.com/LCAS/bacchus_lcas}} (see Figure~\ref{fig:simulation_model}). We also performed tests in a real vineyard located on the Piacenza (Italy) campus of the Università Cattolica del Sacro Cuore. The simulated environment consisted of three vineyard corridors approximately \SI{36}{\metre} long and approximately \SI{2}{\metre} large, characterized by three different vegetative stages: low, medium, and high. The results reported for the simulated environment are thus an average over the three vegetative stages. The real environment was a single vineyard corridor with a length of approximately \SI{40}{\metre} and a width of approximately \SI{2.5}{\metre}, which is one of the typical settings in Italy. The vegetative stage of the real vineyard was comparable to the high vegetative stage of the simulated one. During the tests, we reached a maximum linear speed of \SI{2}{\metre\per\second} in the simulated environment and \SI{1}{\metre\per\second} in the real environment for both in-row and end-row navigation. We mounted the Ouster OS1 LiDAR sensor at an approximate height of \SI{1}{\metre} from the ground.

The navigation system ran on an onboard Shuttle XPC (model DS81L15) equipped with an Intel(R) Core(TM) i7-4790S CPU and 8 GB of RAM. The LiDAR sensors produced messages at a frequency of \SI{10}{\hertz}, and the odometry was published at \SI{50}{\hertz}. All the ROS nodes were capable of keeping up with the \SI{10}{\hertz} frequency of the LiDAR, except for the nodes responsible for clustering and end point detection, which proved to be the bottleneck of the system. Specifically, the node performing clustering with the Nearest end point picking policy operated at a minimum frequency of \SI{9}{\hertz}, while the one using the Line fitting policy ran at a minimum frequency of \SI{5}{\hertz}. Nevertheless, the bottleneck only affected the end-row navigation, which represents a small part of the total path traversed in a vineyard.

\subsection{In-Row Navigation Evaluation}
To evaluate the precision of the In-row navigation module, we measured the robot's displacement from the central row line. This displacement was determined by calculating the absolute distance between the robot's center and the central line of the row. In the simulated environment, we had access to the true robot position, whereas in the real-world test, we relied on the side distance measurements of the In-row algorithm performed via the LiDAR (which has a precision of \SI{\pm 0.01}{\metre}). Evaluating navigation accuracy in real agricultural environments is a challenging and ambiguous task currently addressed by agricultural robotics competitions such as that described in~\cite{bertoglio2021design}. Alternatively, one could utilize an expensive yet highly accurate laser position tracking system, although determining the optimal target trajectory remains a nontrivial problem. In our case, we defined a perfectly row-centered trajectory as the optimal one. However, in both the simulation and the real vineyard, protruding vegetation and branches caused the robot to deviate from the central line, resulting in some average deviation from the center. Table~\ref{tab:inrow_results} presents the outcomes of in-row navigation tests performed in simulation across three rows at varying vegetation stages and in two real vineyard rows.

\begin{table}[htbp]
\centering
\begin{tabular}{l|c|c}
\toprule
\textbf{Measurements} & \textbf{Simulation} & \textbf{Real} \\
\midrule
Mean center displacement & \SI{0.049}{\metre} & \SI{0.372}{\metre} \\
Max center displacement & \SI{0.167}{\metre} & \SI{1.183}{\metre} \\
Mean corridor width & \SI{1.373}{\metre} & \SI{2.142}{\metre} \\
Max corridor width & \SI{2.300}{\metre} & \SI{2.620}{\metre} \\
Min corridor width & \SI{0.740}{\metre} & \SI{1.600}{\metre} \\
\bottomrule
\end{tabular}
\caption{In-row navigation evaluation results.}
\label{tab:inrow_results}
\end{table}

The mean displacement from the central line was \SI{0.049}{\metre} in the simulated environment, whereas in the real vineyard, we observed a mean displacement of \SI{0.372}{\metre}. In both scenarios, the robot successfully avoided protruding branches and never collided with the row sides. Table~\ref{tab:inrow_results} also presents the row width measurements computed from LiDAR scans. The measurements indicate that protruding vegetation causes row width variations, impacting robot centering. In the real scenario, the minimum measurable row width of \SI{1.6}{\metre} was reached, as our LiDAR has a minimum scanning distance of \SI{0.8}{\metre}.

\subsection{Row Ends Detection Evaluation}
To estimate the accuracy of the row ends detection, we computed the Euclidean distance between the true center of row support poles and those detected by our row ends detection system. It is important to note that the assumption that the pole center is always the true row end point is not always valid, as vegetation can cover the pole and protrude outward. In the simulated environment, we computed the instantaneous Euclidean distance from the real pole center to the end point detected by our system during a full turn from one row to the next. We performed measurements for three different vegetative stages. In the real environment, obtaining multiple measurements of the real displacement of the pole center from the robot is laborious and time-consuming. Furthermore, without any absolute positioning system available, the only way to measure it was manually, which introduced measurement errors in the order of centimeters. Therefore, we statically positioned the robot in the middle of a row to detect the two side end points and compared them to manual measurements.

In both the simulated and real scenarios, we compared the two policies explained in section~\ref{sec:endrow_navigation}: Nearest and Line fitting. Table~\ref{tab:endrow_points_results} shows the mean, max, and min distances between the true center poles coordinates and those detected by our system. In the simulated scenario, the Line fitting policy was more accurate with a mean of \SI{0.155}{\metre}. The Nearest policy also showed an acceptable mean distance of \SI{0.205}{\metre} while being less computationally intensive. In the real scenario, the accuracy of both policies was comparable since the difference in the order of centimeters could be attributable to the error of manual measurements. Nonetheless, our row ends detection system performed accurately in both scenarios.

\begin{table}[t]
\centering
\begin{tabular}{l|c|c|c|c}
\toprule
\textbf{Pole distance} & \multicolumn{2}{c}{\textbf{Simulation}} & \multicolumn{2}{c}{\textbf{Real}} \\
\cmidrule(lr){2-3}
\cmidrule(lr){4-5}
\textbf{error} & \textbf{Nearest} & \textbf{Line fitting} & \textbf{Nearest} & \textbf{Line fitting} \\
\midrule
Mean & \SI{0.205}{\metre} & \SI{0.155}{\metre} & \SI{0.23}{\metre} & \SI{0.26}{\metre} \\
Max & \SI{0.540}{\metre} & \SI{0.363}{\metre} & \SI{0.30}{\metre} & \SI{0.32}{\metre} \\
Min & \SI{0.038}{\metre} & \SI{0.013}{\metre} & \SI{0.15}{\metre} & \SI{0.20}{\metre} \\
\bottomrule
\end{tabular}
\caption{Row end points detection evaluation.}
\label{tab:endrow_points_results}
\end{table}

\section{Conclusions}

In this paper, we have presented a simple and efficient map-free LiDAR-based navigation system designed for vineyard applications. Our approach relies on the geometrical structure of the environment and does not require a pre-built map or GNSS measurements. The navigation system is capable of in-row, turn, and end-row navigation and has been tested in both simulated and real vineyards. The results of our experiments indicate that the proposed navigation system achieves accurate and reliable navigation performance, even under challenging vineyard conditions with variations in row spacing and vegetative stages. The system can effectively detect protruding vegetation and adjust the trajectory accordingly, potentially reducing crop damage. The proposed navigation system is simple and cost-effective, relying only on odometry and LiDAR as sources of information, requiring low computational effort. Future work can explore testing with a 2D LiDAR to compare the navigation precision and extend the system's evaluation to other types of line-arranged crops. Additionally, the system could be integrated with a robust semantic obstacle detection algorithm to enhance the navigation system's safety.

\addtolength{\textheight}{-12cm}   





\section*{ACKNOWLEDGMENT}
We are grateful to our colleagues at Università Cattolica del Sacro Cuore (Piacenza, Italy), especially Professor Matteo Gatti, for allowing us to conduct experiments in their vineyard on the university campus. This study was conducted within the Agritech National Research Center, and received partial funding from the European Union Next-GenerationEU (Piano Nazionale di Ripresa e Resilienza (PNRR), missione 4, componente 2, investimento 1.4, D.D. 1032 17/06/2022, CN00000022), the European Union's Digital Europe Programme under grant agreement N.101100622, and the European Union's H2020 grant N.101016577.
\bibliographystyle{IEEEtran}
\bibliography{references}

\begin{thebibliography}{10}
\providecommand{\url}[1]{#1}
\csname url@rmstyle\endcsname
\providecommand{\newblock}{\relax}
\providecommand{\bibinfo}[2]{#2}
\providecommand\BIBentrySTDinterwordspacing{\spaceskip=0pt\relax}
\providecommand\BIBentryALTinterwordstretchfactor{4}
\providecommand\BIBentryALTinterwordspacing{\spaceskip=\fontdimen2\font plus
\BIBentryALTinterwordstretchfactor\fontdimen3\font minus
  \fontdimen4\font\relax}
\providecommand\BIBforeignlanguage[2]{{%
\expandafter\ifx\csname l@#1\endcsname\relax
\typeout{** WARNING: IEEEtran.bst: No hyphenation pattern has been}%
\typeout{** loaded for the language `#1'. Using the pattern for}%
\typeout{** the default language instead.}%
\else
\language=\csname l@#1\endcsname
\fi
#2}}

\bibitem{bertoglio2021digital}
R.~Bertoglio, C.~Corbo, F.~M. Renga, and M.~Matteucci, ``The digital
  agricultural revolution: a bibliometric analysis literature review,''
  \emph{IEEE Access}, vol.~9, pp. 134\,762--134\,782, 2021.

\bibitem{guzman_aut_2016}
R.~Guzm{\'a}n, J.~Ari{\~n}o, R.~Navarro, C.~Lopes, J.~Gra{\c{c}}a, M.~Reyes,
  A.~Barriguinha, and R.~Braga, ``Autonomous hybrid gps/reactive navigation of
  an unmanned ground vehicle for precision viticulture-vinbot,''
  \emph{Intervitis Interfructa Hortitechnica-Technology for wine, juice and
  special crops}, 2016.

\bibitem{santos_towards_2016}
F.~N. Dos~Santos, H.~Sobreira, D.~Campos, R.~Morais, A.~Paulo~Moreira, and
  O.~Contente, ``Towards a reliable robot for steep slope vineyards
  monitoring,'' \emph{Journal of Intelligent \& Robotic Systems}, vol.~83, pp.
  429--444, 2016.

\bibitem{bernad_eval_2018}
P.~Bernad, P.~Lepej, {\v{C}}.~Rozman, K.~Pa{\v{z}}ek, and J.~Rakun, ``An
  evaluation of three different infield navigation algorithms,'' in
  \emph{Agricultural Robots-Fundamentals and Applications}, J.~Zhou and
  B.~Zhang, Eds.\hskip 1em plus 0.5em minus 0.4em\relax IntechOpen, 2019,
  ch.~3.

\bibitem{mas_augm_2021}
F.~Rovira-M{\'a}s, V.~Saiz-Rubio, and A.~Cuenca-Cuenca, ``Augmented perception
  for agricultural robots navigation,'' \emph{IEEE Sensors Journal}, vol.~21,
  no.~10, pp. 11\,712--11\,727, 2020.

\bibitem{mengoli_aut_2020}
D.~Mengoli, R.~Tazzari, and L.~Marconi, ``Autonomous robotic platform for
  precision orchard management: Architecture and software perspective,'' in
  \emph{2020 IEEE International Workshop on Metrology for Agriculture and
  Forestry (MetroAgriFor)}.\hskip 1em plus 0.5em minus 0.4em\relax IEEE, 2020,
  pp. 303--308.

\bibitem{mengoli_robust_2021}
D.~Mengoli, A.~Eusebi, S.~Rossi, R.~Tazzari, and L.~Marconi, ``Robust
  autonomous row-change maneuvers for agricultural robotic platform,'' in
  \emph{2021 IEEE International Workshop on Metrology for Agriculture and
  Forestry (MetroAgriFor)}.\hskip 1em plus 0.5em minus 0.4em\relax IEEE, 2021,
  pp. 390--395.

\bibitem{aghi_local_2020}
D.~Aghi, V.~Mazzia, and M.~Chiaberge, ``Local motion planner for autonomous
  navigation in vineyards with a rgb-d camera-based algorithm and deep learning
  synergy,'' \emph{Machines}, vol.~8, no.~2, p.~27, 2020.

\bibitem{kamaro2021fre}
\BIBentryALTinterwordspacing
J.~Bier, T.~Friedel, J.~Barthel, E.~Bulovas, and L.~Tuschla, ``{BETEIGEUZE -
  KAMARO},'' pp. 17--22, 2022. [Online]. Available:
  \url{https://www.fieldrobot.com/event/wp-content/uploads/2022/02/Proceedings_FRE2021.pdf}
\BIBentrySTDinterwordspacing

\bibitem{odometry_paper}
A.~Mandow, J.~L. Martinez, J.~Morales, J.~L. Blanco, A.~Garcia-Cerezo, and
  J.~Gonzalez, ``Experimental kinematics for wheeled skid-steer mobile
  robots,'' in \emph{2007 IEEE/RSJ international conference on intelligent
  robots and systems}.\hskip 1em plus 0.5em minus 0.4em\relax IEEE, 2007, pp.
  1222--1227.

\bibitem{RusuDoctoralDissertation}
R.~B. Rusu, ``Semantic 3d object maps for everyday manipulation in human living
  environments,'' Ph.D. dissertation, Computer Science department, Technische
  Universitaet Muenchen, Germany, October 2009.

\bibitem{bertoglio2021design}
R.~Bertoglio, G.~Fontana, M.~Matteucci, D.~Facchinetti, M.~Berducat, and
  D.~Boffety, ``On the design of the agri-food competition for robot evaluation
  (acre),'' in \emph{2021 IEEE International Conference on Autonomous Robot
  Systems and Competitions (ICARSC)}.\hskip 1em plus 0.5em minus 0.4em\relax
  IEEE, 2021, pp. 161--166.

\end{thebibliography}

\end{document}